\documentclass[journal]{IEEEtran}

\usepackage{booktabs}
\usepackage{indentfirst}
\usepackage{graphicx}
\usepackage{amsmath}
\usepackage{latexsym}
\usepackage{amssymb}
\usepackage{subfigure}
\usepackage{multirow}
\usepackage{cite}
\usepackage{algorithm}
\usepackage{algorithmic}
\usepackage{color}

\begin{document}
%
\title{Pose Discrepancy Spatial Transformer Based Feature Disentangling for Partial Aspect Angles SAR Target Recognition}
%
%
\author{Zaidao Wen,~\emph{Member,~IEEE},~Jiaxiang Liu, Zhunga Liu,~\emph{Member,~IEEE}, and Quan Pan,~\emph{Member,~IEEE}}

\maketitle

\begin{abstract}
This letter presents a novel framework termed DistSTN for the task of synthetic aperture radar (SAR) automatic target recognition (ATR). In contrast to the conventional SAR ATR algorithms, DistSTN considers a more challenging practical scenario for non-cooperative targets whose aspect angles for training are incomplete and limited in a partial range while those of testing samples are unlimited. To address this issue, instead of learning the pose invariant features, DistSTN newly involves an elaborated feature disentangling model to separate the learned pose factors of a SAR target from the identity ones so that they can independently control the representation process of the target image. To disentangle the explainable pose factors, we develop a pose discrepancy spatial transformer module in DistSTN to characterize the intrinsic transformation between the factors of two different targets with an explicit geometric model. Furthermore, DistSTN develops an amortized inference scheme that enables efficient feature extraction and recognition using an encoder-decoder mechanism. Experimental results with the moving and stationary target acquisition and recognition (MSTAR) benchmark demonstrate the effectiveness of our proposed approach. Compared with the other ATR algorithms, DistSTN can achieve higher recognition accuracy.

%
%
%
%
\end{abstract}


%
\IEEEpeerreviewmaketitle

\section{Introduction}\label{Sec:Introduction}

\PARstart{A}{utomatic} target recognition (ATR) is one of the ultimate goals in the field of synthetic aperture radar (SAR), which has attracted much attention in civil and military reconnaissance and surveillance for years \cite{Novak1997,Keydel1996,El-Darymli2016}. In contrast to optical sensors, the special electromagnetic imaging mechanism of SAR makes the resulted image reconstruction of the specular backscattering of the illuminated target \cite{Niu2020}. Therefore, SAR will actually ``see" some types of physical structures of a target, whose scattering signatures in the resulted SAR image will be highly sensitive to its pose. Crucially, the profile and the position of strong scattering points of the target will vary a lot as the viewing angle of the SAR platform changes, which makes SAR ATR more challenging in comparison with the general recognition task with data of optical sensors.

\par To tackle this issue, an intuitive solution in conventional SAR ATR algorithms is to collect the distributions of strong scattering points of the target in full ($0\sim 360^\circ$) aspect angles uniformly \cite{Potter1997,Jianxiong2011}, which will be subsequently constructed as a feature template in the hope to record all the scattering signatures in different poses. Then the identity of a query test target can be determined in a pairwise class-template matching manner. However, this approach is not robust to many disturbances such as speckle noise and motion ambiguity, which heavily influence its practical recognition performance. To improve the robustness, several hand-crafted image feature extractors such as scale-invariant feature transform (SIFT) focus on some pose invariant features to describe some intrinsic visual signatures of the target. These visual features will be normally more robust and achieve a better generalization performance by choosing a suitable discriminative classifier such as support vector machine (SVM) \cite{Wang2006}. To further improve the discrimination and adaptivity of the features, many machine learning-based algorithms are gradually devoted to SAR ATR \cite{Zhang2012,Wright2009,Dong2016,Dong2015,Wen2018,Bai2019,Chen2016,Deng2017,ding2016convolutional,Song2017,Zhou2018}, among which the most notable model will be the convolutional neural networks (CNN). Distinct from the above feature engineering algorithms, the core idea of CNN-based algorithms to address the pose sensitivity issue is to fit a set of rotating target images as well as their class-labels with a  deep neural network through which every intra-class rotating targets can be mapped to the same class-label. In this way, some label-invariant features are expected to be obtained in a discriminative learning way without concerning their variances in aspect angle, and the recognition performance can be remarkably improved in terms of both accuracy and efficiency. 
\par The success of the above learning models, especially CNN owns to the existence of a large number of training samples to cover sufficient target patterns. It can ensure the distribution consistency between the training and testing ones. However, in practical SAR ATR application, it is intractable to collect sufficient non-cooperative target samples in the different poses for training. In general, only a small number of targets in partial aspect angles can be acquired by reconnaissance. It will thus lead to the so-called out of distribution (o.o.d) classification problem that the distributions of the training and testing samples are slightly different \cite{Geirhos2020}. In this case, the recognition performance of the above SAR ATR algorithms will dramatically decrease due to a higher requirement of the generalization ability \cite{Geirhos2020}. An intuitive way to overcome this problem in computer vision society would be manually generating pseudo targets in full angles for data augmentation and distribution completion and alignment. We, however, empirically found that this trick is invalid for SAR data, which might be due to ambiguous artifacts caused by nonlinear pixel interpolation. The interpolated pixels will not correctly recover the actual physical scattering signature of the corresponding real target so that these generated pseudo samples do not explicitly provide more discrimination information than the original ones \cite{Zhou2018,Pei2018,Schmidt2012}.

\par Motivated by the above practical o.o.d scenario, this paper develops a pose discrepancy spatial transformer based feature disentangling framework (DistSTN) for partial aspect angles SAR target recognition (PAA-ATR). Instead of learning the pose invariant features, DistSTN newly involves an elaborated feature disentangling model to separate the learned pose factors of a SAR target from the identity ones so that they can independently control the representation process of the target image for better generalization ability.  To disentangle the explainable factors, a pose discrepancy spatial transformer module is developed in DistSTN. It aims to characterize the intrinsic transformation between the factors of two different targets with an explicit geometric model induced regularization. Furthermore, DistSTN develops an amortized inference scheme that enables efficient feature extraction and recognition using an encoder-decoder mechanism. Experimental results on the MSTAR benchmark demonstrate that our framework achieves better recognition accuracy in the PAA-ATR task. The rest of this paper is organized as follows: Section \ref{Sec:framework} proposes the main framework; Section \ref{Sec:Experiment} describes our experiments to validate performance and Section \ref{Sec:Conclusion} summarizes our work and suggests future directions.  
\section{Framework Presentation}\label{Sec:framework}
In this section, we will first formulate the problem of PAA-ATR and analysis some insight viewpoints to shed light on our solution. Next, we will develop a spatial transformer-based feature disentanglement model to address this task. Finally, we will propose an encoder-decoder architecture for amortized inference and model learning.
\subsection{Problem Formulation of PAA-ATR and Insight}
\par Let $\{\mathbf{x}_i,\mathbf{y}_i\}_{i=1}^{N_c}$ be $N_c$ labeled SAR targets drawn from c-th class, where $\mathbf{x}_i$ and $\mathbf{y}_i$ stand for the SAR target image and its identity label vector, respectively, and $C$ is the total number of target categories. The ultimate goal of ATR is to predict the label vector of a new query sample according to these training data. In general SAR ATR algorithms, it is implicitly assumed that the aspect angle of the training and testing samples should identically and uniformly reside in the range of $0^{\circ}\sim 360^{\circ}$, which is intractable for non-cooperative targets. Alternatively, this paper considers a more difficult but practical task termed PAA-ATR. It assumes that the aspect angles of the training samples from at least one class are incomplete and limited in a partial range of $0^{\circ}\sim 180^{\circ}$, while those of testing samples are in $0^{\circ}\sim 360^{\circ}$, yielding an o.o.d scenario.
\par  Due to the electromagnetic imaging mechanism of the SAR sensor, the scattering appearance of an illuminated target will be severely sensitive to the relative pose between the target and the sensor. As a result, the difference between the training and testing samples in PAA-ATR will be aggravated in comparison with the general ATR task. The latent factors accounting for the pose and identity are entangled in the image domain. To alleviate this challenge, the conventional algorithms will exploit some physical-driven or geometric-guided methods to design some rotation-invariant features in a hand-crafted way. Alternatively, the learning-based model such as CNN will train a deep network mapping every intra-class targets with different poses into the same label vector in a supervised learning way. Through this way, the intermediate features will only account for the identity label and ignore the intra-class variances, and it can achieve much better performance only assuming the distribution consistency between the training and testing samples. Since the general CNN involves no explicit spatial geometric transformation, it can, however, only achieve local spatial pose invariance by introducing a deep hierarchy of max-pooling and convolutions layers. Consequently, its generalization ability for PAA-ATR is weak.
\subsection{The Framework for Feature Disentangling}
According to the above analysis, the critical issue of the CNN-based model is the lack of rotation awareness so that the model does not understand the physical and semantic concept of the target rotation. In this sense, it is intractable for a CNN-based model to generalize the training rotational pattern to the unseen ones in the testing phase in PAA-ATR. To tackle this issue, our core idea alternatively focuses on equivariant feature disentanglement instead of pursuing the pose-invariant features for discrimination. In contrast to the general CNN model for discriminative learning, we will develop a generative feature learning model containing a geometric transformation module. It aims to explicitly characterize and disentangle the pose features from the identity ones. Through this separation, it is expected that the discrimination of the rest identity ones will be enhanced without being influenced.
\par To this end, let $\mathbf{r}$ and $\mathbf{f}$ be the latent pose (relative to the radar) and identity factors, respectively. Considering a generative learning model, a SAR target image is modeled and represented by $\mathbf{f}$ and $\mathbf{r}$ through a nonlinear parametric function as $\mathbf{x}\gets g_{\Theta}(\mathbf{f,r})$, where $\Theta$ contains the model parameters controlling the unknown intricate SAR imaging process. Based on this generative model, the task of feature disentanglement is an inverse problem of extracting $\mathbf{r}$ and $\mathbf{f}$ from $\mathbf{x}$, which can be usually addressed by maximizing a posterior (MAP) estimator given by:
\begin{equation}\label{Equ:Fea_Disentangle}
	\min_{\mathbf{f,r},\Theta} \Omega_f(\mathbf{f})+\Omega_r(\mathbf{r}),~\mathrm{s.t.}~\mathbf{x}=g_{\Theta}(\mathbf{f,r})+\epsilon
\end{equation} 
where $\Omega_f(\cdot)$ and $\Omega_r(\cdot)$ are two elaborated regularization functions on $\mathbf{f}$ and $\mathbf{r}$ to encode our prior preference for disentangling, and $\epsilon$ measures the representation error. Note from Eq.\eqref{Equ:Fea_Disentangle} that it originates from the idea of independent component analysis (ICA) for source separation \cite{Bell1995}. Therefore, the critical issue in Eq. \eqref{Equ:Fea_Disentangle} is to design two regularization functions for effective disentanglement and develop an efficient feature inference process to solve the optimization.
\par For the first regularizer $\Omega_f(\cdot)$, it can be designed as the following task-induced function \eqref{Equ:Discri_Regularizer} with a supervised analysis prior \cite{Elad2007}, which will force $\mathbf{f}$ to contain sufficient discriminative information for correct recognition.
\begin{equation}\label{Equ:Discri_Regularizer}
		 \vspace{-0.05cm}
\Omega_f(\mathbf{f})\equiv-\log p(\mathbf{y|f})=-\log \Pi_{c=1}^C W(\mathbf{f})_c^{\mathbb{I}(y_c=c)}
\end{equation} 
where $-\log p(\mathbf{y|f})$ is the negative log-likelihood function of $\mathbf{f}$ given $\mathbf{y}$ induced from the categorical distribution,  $\mathbb{I}$ is the indicator function, $W(\mathbf{f})$ is a simple affine transformation of $\mathbf{f}$ followed by a softmax function, and the subscript $c$ represents the corresponding value in c-th index.
\par
\begin{figure*}[h]
	\centering
	\subfigure[{Overall architecture for model training}]{ \label{fig:Framework} 
		\includegraphics[width=0.53\textwidth]{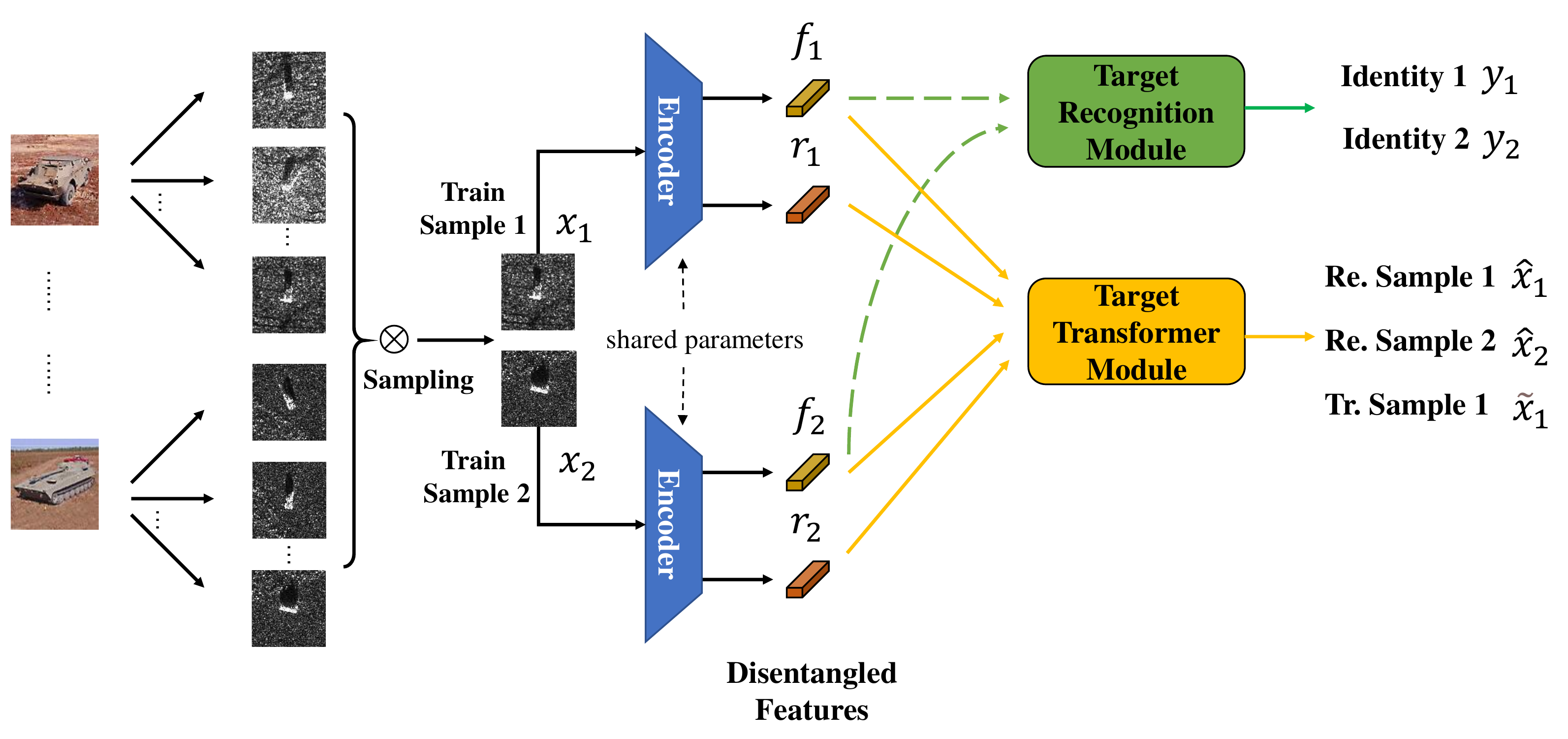}}
	\subfigure[Illustration of feature disentanglement via pose discrepancy STN]{ \label{fig:targettransformermodule} 
		\includegraphics[width=0.45\textwidth]{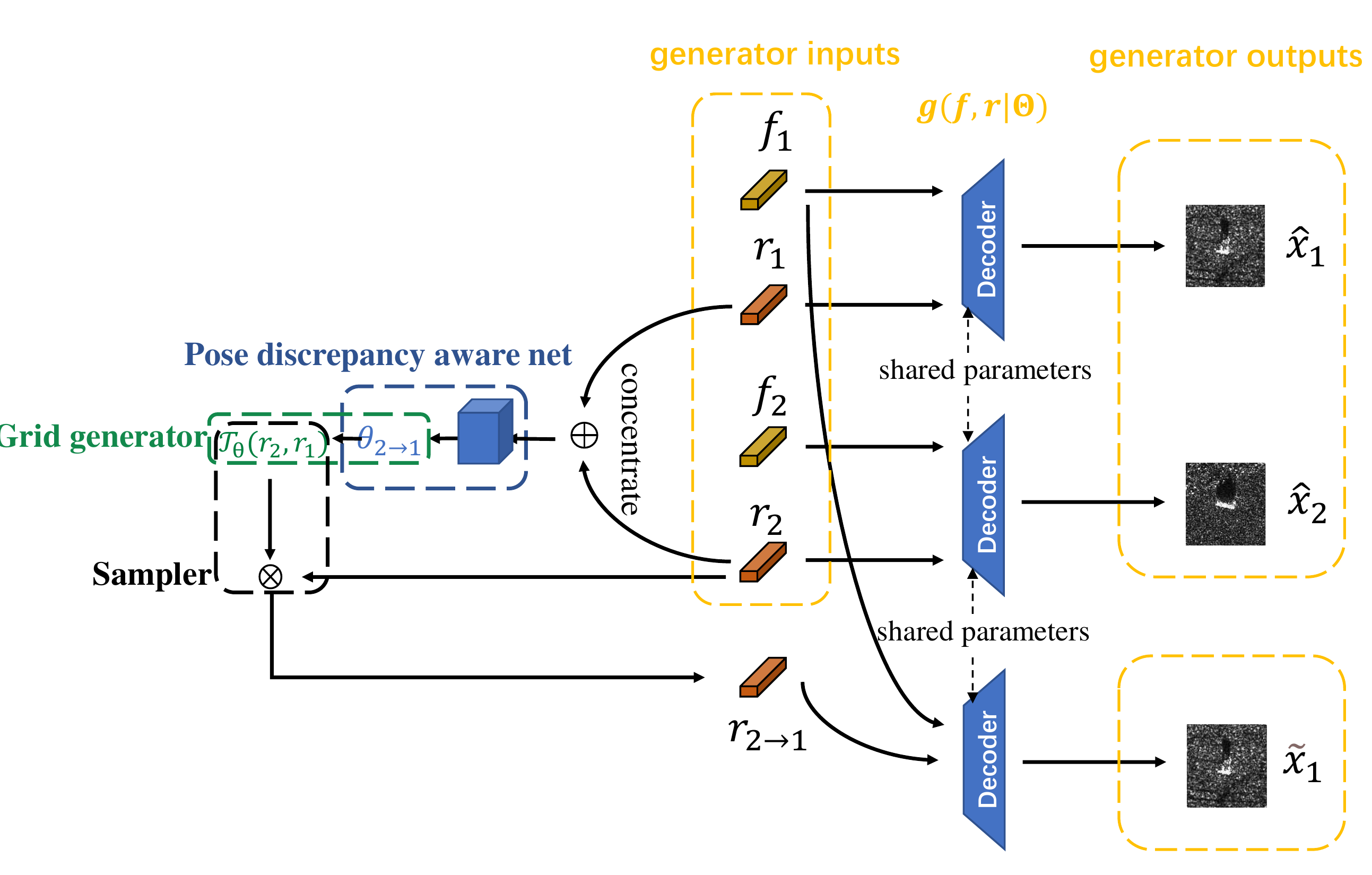}}
	\caption{The framework of the proposed feature disentangling model.}
	\label{fig:overall_architecture}
\end{figure*}

 For the second regularizer $\Omega_r(\cdot)$, it initially attempts to characterize the factors accounting for the target pose relative to the sensor, including azimuth angle, depression angle, and some other positional factors of the sensor. Nevertheless, it is intractable to labeling the entire factors explicitly and exactly for all training targets. Thus, we cannot exploit the above strategy to model it in a discriminative way. Alternatively, we propose a novel self-supervised task of target cross-transformation to model the pairwise pose discrepancy between two targets with an explicit geometric model. Formally, let $\mathbf{r}_i$ and  $\mathbf{r}_j$ be the pose factors of $\mathbf{x}_i$ and $\mathbf{x}_j$, respectively. If they can indeed capture the entire pose information, there will be a geometrically explicit operator $\mathcal{T}_{\theta_{i\rightarrow j}}$ warping $\mathbf{r}_i$ to $\mathbf{r}_j$ and vice versa. More importantly, the parameters $\theta_{i\rightarrow j}$ will have a clear physical meaning to measure the pose discrepancy between two targets without being influenced by the other shared sensor factors. According to the 2D rigid-body geometric transformation model \cite{Ma}, the warping function $\mathcal{T}_{\theta_{i\rightarrow j}}$ is essentially an affine transformation of the 2D coordinates of the input feature maps, followed by sampling and interpolation processes that is also exploited in the spatial transformer network (STN) \cite{Jaderberg2015}. If $\theta_{i\rightarrow j}$ can be correctly estimated and assigned, $\mathcal{T}_{\theta_{i\rightarrow j}}(\mathbf{r}_i)$ will be equal to $\mathbf{r}_j$. {Therefore, $\mathcal{T}_{\theta_{i\rightarrow j}}(\mathbf{r}_i)$ can be further exploited to represent $\mathbf{x}_j$ in conjunction with $g_{\Theta}$ and $\mathbf{f}_j$.} In this sense, we can exploit a pose discrepancy aware network to estimate the parameters as $\theta_{i\rightarrow j}=P(\mathbf{r}_i,\mathbf{r}_j|\varphi)$, where $\varphi$ contains its parameters. {$P$ and $\mathcal{T}_{\theta_{i\rightarrow j}}$ will constitute the designed pose discrepancy STN illustrated in Fig. \ref{fig:targettransformermodule}.} According to this model, $\Omega_{r}$ will be designed as \eqref{Equ:pose_regulazi} to measure the error of representing $\mathbf{x}_j$ with $\mathbf{f}_j$ and $\mathcal{T}_{\theta_{i\rightarrow j}}(\mathbf{r}_i)$ without external supervised pose information.
\begin{equation}\label{Equ:pose_regulazi}
	\Omega_{r}(\mathbf{r}_i)=\ell(\mathbf{x}_j,g_{\Theta}(\mathbf{f}_j,\mathcal{T}_{\theta_{i\rightarrow j}}(\mathbf{r}_i))),~\mathrm{s.t.}~\mathbf{x}_i=g_{\Theta}(\mathbf{f}_i,\mathbf{r}_i)+\epsilon_i
\end{equation}
\par {It should be worth noting that the main purpose of pose discrepancy STN is not to generate a high-quality SAR image to simulate its special imaging mechanism, but to impose an explicit model-induced learning bias on the latent $\mathbf{r}$ for disentanglement.}
\subsection{Amortized Inference and Overall Architecture}
\par In the above subsection, we have developed a novel model \eqref{Equ:Fea_Disentangle} for identity and pose features disentanglement. We have elaborated two regularization functions \eqref{Equ:Discri_Regularizer} and \eqref{Equ:pose_regulazi} to inject the identity and pose information into the features $\mathbf{f}$ and $\mathbf{r}$, respectively. More specifically, Eq. \eqref{Equ:Discri_Regularizer} exploits a task-induced regularization on each $\mathbf{f}$ to make it more discriminated while Eq. \eqref{Equ:pose_regulazi} involves a geometric transformation model to characterize the pose discrepancy between two targets. However, directly solving the inverse problem \eqref{Equ:Fea_Disentangle} with a general optimization algorithm is intractable and time-consuming. Inspiring from the recent amortized inference \cite{Bell1995} and our previous research \cite{Wen2017a,Wen2018}, we will utilize the encoder-decoder architecture for feature inference and parameters learning in an end-to-end learning pipeline. To this end, we will design an encoder $G$ parameterized by the three-layers CNNs in the hope to directly output the estimated feature maps as $\mathbf{f,r}= G(\mathbf{x}|\Psi)$, where $\Psi$ contains the parameters of $G$ to be learned. Through this way, the identity and pose features in terms of the model \eqref{Equ:Fea_Disentangle} can be efficiently obtained with low computational complexity. The pose discrepancy-aware network is a three-layer fully connected network whose hidden unit numbers are 60, 30, and 6, respectively. The overall architecture termed DistSTN is illustrated in Fig. \ref{fig:Framework}, and the final optimization problem is summarized as:
\begin{equation}\label{Equ:DistSTN}\small			 \vspace{-0.05cm}
\begin{split}
\min \sum_{i,j,i\neq j}\Omega(\mathbf{f}_i)+\alpha\ell(\mathbf{x}_j,g_{\Theta}(\mathbf{f}_j,\mathcal{T}_{\theta_{i\rightarrow j}}(\mathbf{r}_i)))+\beta\ell(\mathbf{x}_i,g_{\Theta}(\mathbf{f}_i,\mathbf{r}_i))\\
+\beta\ell(\mathbf{x}_j,g_{\Theta}(\mathbf{f}_j,\mathbf{r}_j)),~\mathrm{s.t.}~\theta_{i\rightarrow j}=P(\mathbf{r}_i,\mathbf{r}_j|\varphi),~\mathbf{f}_i,\mathbf{r}_i= G(\mathbf{x}_i|\Psi)
\end{split}
\end{equation} 
where $\alpha$ and $\beta$ are two hyper-parameters for balance. From Fig. \ref{fig:Framework}, DistSTN is a double-input CNN which allows taking two targets from arbitrary classes. It follows that we can generate at most $\mathcal{A}_{N}^2$ target pairs for model learning, where $\mathcal{A}$ is the permutation operator and $N$ counts the total number of training samples. In this regard, DistSTN will be more appropriate for learning with limited training samples. In the testing phase, we can simply remove the target transformer module from DistSTN and feed the query sample into the encoder-target recognition module to output its identity label.

\section{Experiments}\label{Sec:Experiment}
\par In this section, we carried out several experiments on the MSTAR database to validate the performance of the proposed DistSTN for PAA-ATR. The parameters in the networks are initially in a default way without pre-training. We exploit the weight decay regularization on the parameters in with rate $0.004$, except $\varphi$ in $P$. The optimizer of DistSTN is chosen as the stochastic gradient descent (SGD) with a constant learning rate of $0.001$ and a momentum rate of $0.9$\footnote{It is empirically found that using the Adam optimizer can obviously achieve a much higher recognition accuracy for the most compared algorithms.}. Two hyper-parameters $\alpha$ and $\beta $ are determined via cross-validation according to grid search. The loss function $\ell$ is chosen as the mean absolute error. We exploit the early-stopping trick to control the training procedure. The model achieving the best performance on a validate set will be restored for testing. We conduct all the experiments on a workstation with a single RTX 2070-Super GPU five times using TensorFlow 2.3 library.

\subsection{Database and Comparison Algorithms Introduction}
\subsubsection{Database} The MSTAR database was collected by the Sandia National Laboratory using a Twin Otter SAR sensor operating at X-band. It comprises about ten types of military ground target images taken at multiple depression angles and $0^{\circ}\sim360^\circ$ aspect angles with approximately $5^\circ$ interval. We crop all input amplitude images to obtain the central $88\times 88$ target chip to get rid of the impact from the surrounding background cluttering. According to the normal setting in other SAR ATR algorithms \cite{Chen2016}, the targets taken at $17^\circ$ and $15^\circ$ will be used for training and testing respectively. {In particular, to verify the performance of DistSTN on PAA-ATR, the detailed training and testing information different from the normal setting is summarized in Table \ref{Tab_PAA-ATR}. The training set comprises cooperative and non-cooperative classes to simulate the practical situation. In order to keep number of samples in the cooperative and non-cooperative classes balanced, only a half of cooperative samples in each class will be used for training. For all testing samples, their aspect angles are unlimited.} Considering the limitation of computation memory and time cost, we randomly shuffle all training samples twice to generate two training sets from which $\mathbf{x}_i$ and $\mathbf{x}_j$ will be jointly sampled for model learning. 
\begin{table}[]\centering
	\caption{{Information of Training and Testing Setting for PAA-ATR}}\label{Tab_PAA-ATR}
	\begin{tabular}{|l|c|c|c|}
		\hline
		\multicolumn{2}{|l|}{}                      & Aspect Angle     & Depression Angle \\ \hline
		\multirow{2}{*}{Training} & Coop.     & $0\sim 360^{\circ}$            & $17^{\circ}$               \\ \cline{2-4} 
		& Non-Coop. & $0\sim 180^{\circ}$ or  $180^\circ\sim 360^{\circ}$ &  $17^{\circ}$               \\ \hline
		\multicolumn{2}{|l|}{Testing}        & $0\sim 360^{\circ}$ &  $15^{\circ}$              \\ \hline
	\end{tabular}
\end{table}
%
%

\subsubsection{Algorithms} We will compare several existing SAR ATR algorithms to demonstrate the effectiveness and superiority of our proposal, including aforementioned support vector machine (SVM) with a Gaussian kernel, SRC \cite{Wright2009} and A-ConvNets \cite{Chen2016}. Additionally, STN is a designed free module to handle the rotations among inputs \cite{Jaderberg2015}. For comparison, the STN module will be inserted into the A-ConvNets, namely A-ConvNets+STN. We also exploit data augmentation trick of generating some pseudo targets to form a full aspect angle training set. It will be used to train A-ConvNet, yielding another variant termed A-ConvNet*. Finally, ResNet-50 and EfficientNet, two state-of-the-art architectures for image classification will be also compared \cite{He2016,Tan2019}.

\begin{figure}
	\centering
	\includegraphics[width=0.47\textwidth]{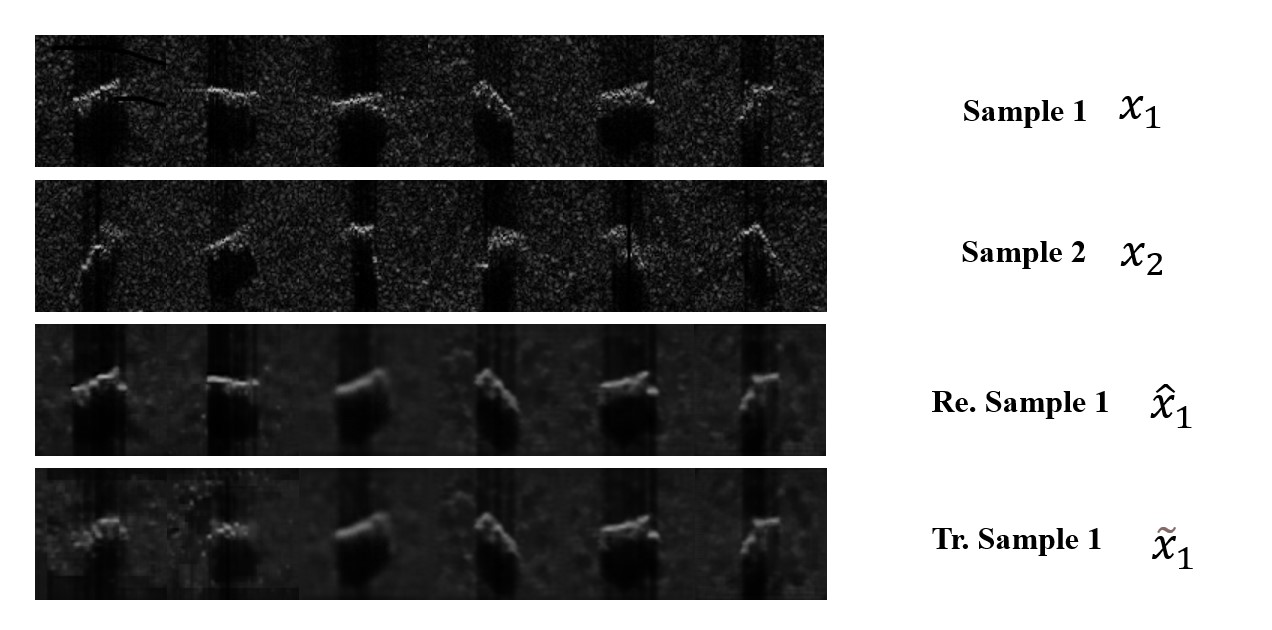}
	\caption{Illustration of cross-reconstruction of DistSTN.}
	\label{fig:reconstruction}
	\vspace{-0.5cm}
\end{figure}

\subsection{Validation of Target Transformation}
Considering the proposed DistSTN, the most notable contribution is to explore an equivariant feature disentanglement model to address the task of PAA-ATR. In order to capture the pose information of a SAR target in the learned feature maps $\mathbf{r}$, we develop a pose discrepancy STN in the hope to wrap the pose features of a target $\mathbf{x}_2$ into those of another target $\mathbf{x}_1$ with the explainable geometric transformation model. If the obtained $\mathbf{r}$ can indeed contain sufficient pose information via DistSTN, the transformed pose feature maps will be able to reconstruct the $\mathbf{x}_1$. As a result, we will design an experiment to validate the effectiveness of the proposed pose discrepancy STN according to its cross-reconstruction result. To this end, we will visualize some resulted reconstruction images shown in Fig. \ref {fig:reconstruction}. The images in the first and second row are initial inputs $\mathbf{x}_1$ and $\mathbf{x}_2$. The third-row illustrates the corresponding reconstruction results of $\mathbf{x}_1$ using $\mathbf{f}_1$ and $\mathbf{r}_1$. The last row depicts the corresponding cross-reconstruction results using the identity features $\mathbf{f}_1$ and $\mathcal{T}_{\theta_{2\rightarrow 1}}(\mathbf{r}_2)$. At the first sight of the results, we can see that the reconstruction results $\hat{\mathbf{x}_1}$ are very similar to the cross-reconstruction ones $\tilde{\mathbf{x}_1}$. Both of these resulted images can be considered as the denoised and smoothed version of the original inputs with the same pose (the target orientation), though the pose discrepancy between $\mathbf{x}_1$ and $\mathbf{x}_2$ are obvious. Therefore, these results can clearly demonstrate the effectiveness of the proposed model for pose feature disentanglement. 
\subsection{Validation of PAA-ATR}
\begin{table}[t]
	\centering
	\caption{{Recognition accuracy of different algorithms}}
	
	\begin{tabular}{c|c|c|c}
		\toprule
		{Methods} & 5 Non-Coo. & 9 Non-Coo.& 10 Non-Coo.\\
		\midrule                               
		{SVM} & 18.26 & 18.26 &18.26\\
		{SRC} & 62.98 & 63.89 &65.31\\
		{A-ConvNets} & 67.29 & 66.03&64.98 \\
		{A-ConvNets$^*$} & 65.35 & 64.03 &63.75\\
		{A-ConvNets+STN} & 68.68 & 67.70 &67.87\\
		{ResNet-50} & 64.98 & 65.42 &66.86\\
				{EfficientNet} & 59.32 & 56.36 &59.88\\
		\textbf{DistSTN} & \textbf{70.72} & \textbf{68.69} &\textbf{69.16}\\
		\bottomrule
		
	\end{tabular}	\label{Tab:Acc}
	\vspace{-0.5cm}	
\end{table}
\par Finally, DistSTN will be compared with the other algorithms on the task PAA-ATR. To validate the performance of discriminative feature disentanglement, different number of non-cooperative class will be considered, including 5 (half), 9 (only one cooperative class) and 10 (all non-cooperative class). The comparison result are summarized in Table \ref{Tab:Acc}. From the results, DistSTN can achieve the highest accuracy among all comparison algorithms for all three settings, which clearly verifies the superior discrimination and generalization ability of disentangled features. The accuracy obtained from SVM is particularly low, which implies this method failing to cope with the unseen angles. A-ConvNets$^*$ trained with the manually generating pseudo targets will be worse than its original counterpart, namely A-ConvNet. This result reflects that the trick used in RGB image classification may be unsuitable for SAR ATR anymore because of different imaging mechanisms. We can conclude from the consequences of A-ConvNets+STN, STN module can indeed improve the recognition performance of A-ConvNet by involving a self-rotational transformation of feature maps. {However, due to different mechanisms and motivations, the generalization ability of STN is still weak than our proposed DistSTN. It is still less efficiency for STN to address such o.o.d classification task.}

\section{Conclusion}\label{Sec:Conclusion}
In this letter, we present an efficient framework termed DistSTN to address the challenging task of PAA-ATR for non-cooperative targets. Instead of pursuing the pose invariant features in the conventional algorithms, DistSTN newly exploits a feature disentangling strategy to separate the pose factors of a target from the identity ones so that they can independently control the representation process of the target. Experimental results 
demonstrate the superiority of our proposed model on PAA-ATR, which achieves higher recognition accuracy compared to the other ATR algorithms. Future research will consider the potential contribution of those cooperative training samples for knowledge transfer.

\bibliographystyle{IEEEtran}
\bibliography{IEEEabrv,refs}
\end{document}